# A study on using image based machine learning methods to develop the surrogate models of stamp forming simulations

Haosu Zhou
Dyson School of Design Engineering
Imperial College London
Exhibition Road, London, SW7 2DB, UK
h.zhou19@imperial.ac.uk

Qingfeng Xu
School of Computing and Information Systems
University of Melbourne
VIC 3010 Australia

Nan Li[1*]
Dyson School of Design Engineering
Imperial College London
Exhibition Road, London, SW7 2DB, UK
n.li09@imperial.ac.uk

**Abstract**

In the design optimization of metal forming, it is increasingly significant to use surrogate models to analyse the finite element analysis (FEA) simulations. However, traditional surrogate models using scalar based machine learning methods (SBMLMs) fall in short of accuracy and generalizability. This is because SBMLMs fail to harness the location information of the simulations. To overcome these shortcomings, image based machine learning methods (IBMLMs) are leveraged in this paper. The underlying theory of location information, which supports the advantages of IBMLM, is qualitatively interpreted. Based on this theory, a Res-SE-U-Net IBMLM surrogate model is developed and compared with a multi-layer perceptron (MLP) as a referencing SBMLM surrogate model. It is demonstrated that the IBMLM model is advantageous over the MLP SBMLM model in accuracy, generalizability, robustness, and informativeness. This paper presents a promising methodology of leveraging IBMLMs in surrogate models to make maximum use of info from FEA results. Future prospective studies that inspired by this paper are also discussed.

## 1. Introduction

In the design of metal forming processes, design candidates that defined by design parameters are evaluated by numerical simulations, such as finite element analysis (FEA) (K. Bathe, 2014; N. Li, J. Lin, T. A. Dean, D. Dry, D. Balint, 2014). The simulation results are then used to optimize the design candidates in design optimization. However, these simulations are computation intensive and time consuming, which makes trial-and-error design optimization impractical. To facilitate the optimization, an effectual method is constructing surrogate models that trained on a limited number of simulation results (J. Lu, J. Tong, D. W. Apley, Z. Zhan, W. Chen, 2020). In surrogate models, the design parameters of the design candidates are regarded as inputs while the corresponding simulation results are regarded as outputs. The surrogate models learn the mapping relationship from the inputs to the outputs, which can fast predict metal forming simulation results in the design optimization. To construct surrogate models, machine learning methods have been widely used.

Traditionally, the scalar based machine learning method (SBMLM) is commonly used, which downscale both the design parameters and the simulation results to scalar based indicators. In recent years, SBMLMs have been widely applied to design optimization tasks (H. Wang, L. Chen, E, Li, 2018; F. Gao, S. Ren, C. Lin, Y. Bai, W. Wang, 2018; Y. Shi, Z. Lu, J. Zhou, E. Zio, 2020; T. Hart-Rawung, J. Buhl, M. Bambach, 2020). Despite of these impacts, SBMLMs usually suffer from several shortcomings. Firstly, SBMLM surrogate models generally fail to extract a bijective mapping relationship from inputs to outputs. This makes the models not just inaccurate, but rather non-generalizable. Secondly, SBMLMs can hardly be applied to scenarios where the location information of the simulation results is of vast importance. This information cannot be represented hard-coded scalar indicators. Typical examples of these scenarios include thinning distributions analysis, punch impact lines analysis, scratching on class A surfaces, microstructures evolution and damage (J. Kuhn, M. Schneider, P. Sonnweber-Ribic, T. Böhlke, 2020). These two shortcomings are mostly because of the information losses caused by the downscaling process.

---

* Address all correspondence to this author

To overcome these shortcomings, a representation that superior to scalars is required to mitigate the information losses. This leads to the image based machine learning method (IBMLM). Herein, both the design parameters and simulation results are represented as images, which reserves their full spatial information. Based on these images, a bijective mapping relationship can be extracted using IBMLMs, which is more accurate and more generalizable compared with that extracted using SBMLMs. Though IBMLMs are still at an early stage of development, they have been investigated in several studies. Early studies of IBMLMs mainly focused on fluid dynamics, since the image-like nature of Eulerian meshing in most fluid-dynamics simulations are suitable for using IBMLMs (M. P. Brenner, J. D. Eldredge, J. B. Freund, 2019; S. Bhatnagar, Y. Afshar, S. Pan, K. Duraisamy, S. Kaushik, 2019; J. Kim, C. Lee, 2020). Recently, IBMLMs are being extended to special scenarios of solid mechanics, such as sheet forming simulations (J. Wang, S. Das, R. Rai, C. Zhou, 2018; L. Liang, M. Liu, C. Martin, W. Sun, 2018; Z. Nie, H. Jiang, L. B. Kara, 2020; K. Ren, Y. Chew, Y. F. Zhang, J. Y. H. Fuh, G. J. Bi, 2020). This is because that the inputs and outputs in these special scenarios can be represented by 2D images. To extend IBMLMs to more general scenarios of solid mechanics, further studies are under investigation (S. Banga, H. Gehani, S. Bhilare, S. J. Patel, L. B. Kara, 2018; T. Spruegel, T. Schröppel, S. Wartzack, 2017; C. Diez, 2020; D. Kracker, J. Garcke, A. Schumacher, P. Schwanitz, 2020).

In this article, the theoretical feasibility and practical advantages of IBMLMs are studied. The objective of this article is to qualitatively analyse the theoretical feasibility of IBMLMs and validate their practical advantages. In section 2, the pipelines of SBMLMs and IBMLMs are discussed in detail. To demonstrate the advantage of location information of IBMLMs, a simplistic, half-symmetric model is analysed. Specific machine learning methods in SBMLMs and IBMLMs are also discussed. In section 3, the details of the dataset and the machine learning models used in this study are discussed. The dataset of a quarter metal forming model is established, which is then be divided into a training set and a test set. The architectures and parameters of two machine learning models are then be stated. In section 4, the results of the SBMLM model and the IBMLM model will be compared and discussed from three aspects, including interpolation performance, extrapolation performance, and information advantages. In section 5, conclusions about IBMLMs will be stated; the outlook of IBMLMs will then be demonstrated.

**2. Theory & Methodology**

IBMLMs have been separately discussed and investigated in multiple realms. In the realm of fluid dynamics, a convolutional neural network (CNN) model was developed to predict steady flows in a representative, early-stage study (X. Guo, W. Li, F. Iorio, 2016). In the realm of topology optimization, deep neural networks have been applied to learn the structural features of optimization results (S. Oh, Y. Jung, S. Kim, I. Lee, N. Kang, 2018; S. Rawat, M.-H. H. Shen, 2019). In the realm of solid mechanics, a multi-channel residual CNN, namely StressNet, was developed to predict the stress fields of multiple 2D cantilever structures under varying boundary conditions (Z. Nie, 2020). Based on StressNet, normalization modules were leveraged to further improve the accuracy of DNNs in predicting the stress concentrations in heterogeneous media, such as composite materials (H. Feng, P. Prabhakar, 2020).

It is noteworthy that in the studies discussed above, the geometries and simulation results are compatible with IBMLMs since they are naturally represented by pixels or voxels. To extend IBMLMs to the realm of nonlinear solid mechanics, complex 3D geometries and strong nonlinearities are the major obstacles. Several studies have attempted to apply CNNs to crash simulations (Y. Li, H. Wang, W. Shuai, H. Zhang, Y. Peng, 2018; Y. Li, H. Wang, K. Mo, T. Zeng, 2018). However, these studies were limited to reconstruction tasks, while could not be used for fast predictions.

Despite these obstacles, it is promising to apply IBMLMs to several special scenarios of nonlinear solid mechanics. One of these scenarios is forming. This is because the geometries in forming processes can be projected to 2D images from top view, while the simulations here can also be plotted onto undeformed blank configurations as 2D images. Several pioneering studies have been executed upon this topic. J. Pfrommer and C. Zimmerling originally applied IBMLMs to predict the simulation images of a composite forming process (2018). By counting in more elements of simulation results, IBMLMs outperformed SBMLMs. This original method was further extended to handle varying geometries using pattern recognition or CNNs (C. Zimmerling, D. Dörr, F. Henning, L. Kärger, 2019; C. Zimmerling, D. Trippe, B. Fengler, L. Kärger, 2019). In a following article, these progress about IBMLM in forming were summarized (C. Zimmerling, C. Poppe, L. Kärger, 2020a). Recently, these developed IBMLM models were combined with reinforcement learning to optimize manufacturing parameters (C. Zimmerling, C. Poppe, L. Kärger, 2020ba).

Based on these studies, it is promising to further develop IBMLMs for the fast prediction and optimization of forming processes. However, most of the studies regarded IBMLMs as black boxes. The underlying theory to the advantage of IBMLM has barely been discussed. To fill this gap, in the present paper, the assumption of location information, which makes IBMLMs advantageous, will be raised and verified. The verification will be fulfilled using a half-symmetric metal forming model.

To elaborate, the pipeline of metal forming simulations and surrogate models is presented in Fig 1. At the beginning, scalar based design parameters are defined by engineers and then modelled by computer-aided design (CAD) programs. This leads to image based design parameters, including projected geometry images, distribution images of binder forces and thickness. These CAD models are processed through FEA solvers to acquire image based simulation results, which include multiple physical fields. To extract the mapping relationship from the design parameters to the simulation results, surrogate models are required. Traditionally, due to the lack of advanced representations and modelling tools, the simulation images are downscaled to several critical scalar indicators such as maximum strain. As shown by Fig 1 (a), this downscaling leads to SBMLMs which bridge scalar based design parameters and simulation results. It is obvious that in SBMLMs, the location information of both design parameters and simulation results are discarded. This information loss is assumed to be responsible for the inaccuracy and non-generalizability of SBMLMs. To mitigate this information loss, IBMLMs are thus developed as shown in Fig 1 (b), which directly bridge the images of both design parameters and simulation results. With location information included, IBMLMs are promising to be more accurate and generalizable.

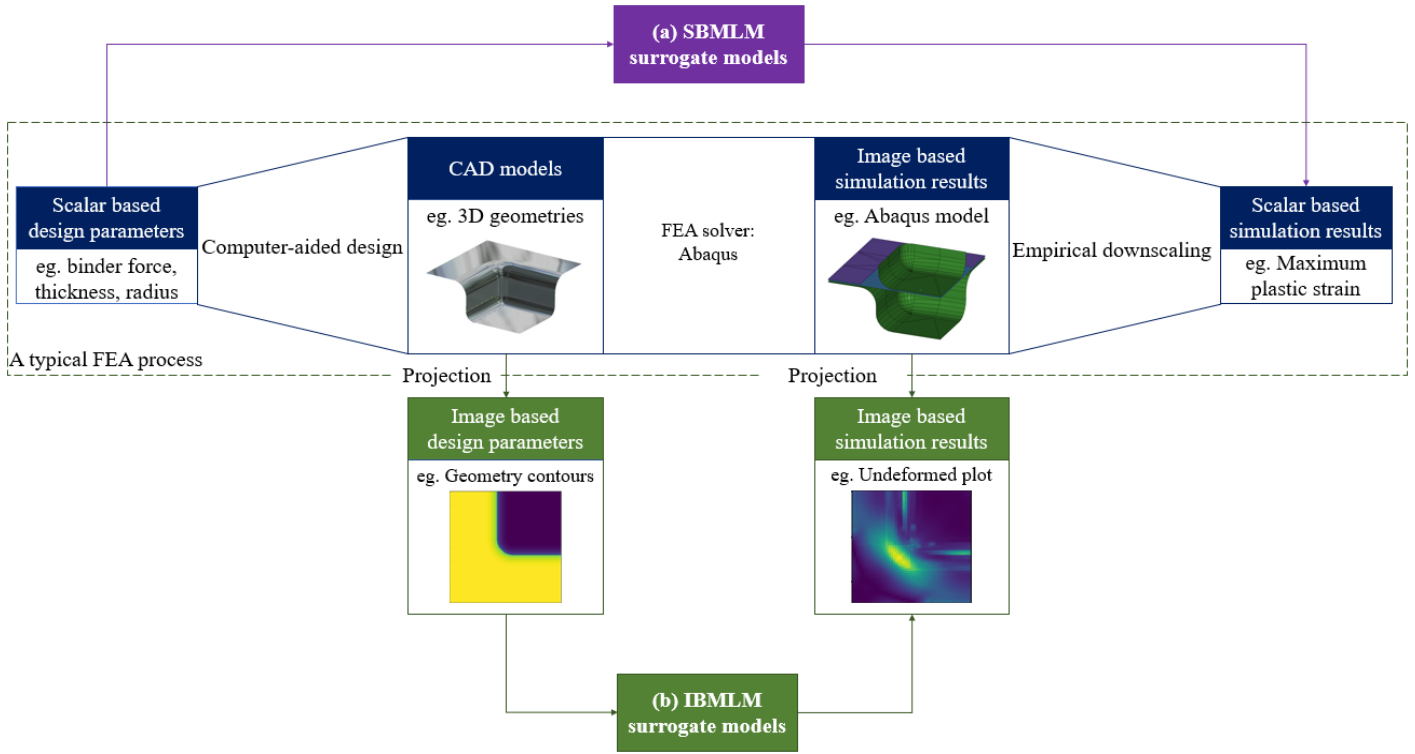

Fig 1. Pipelines of surrogate models: (a) SBMLM surrogate models (b) IBMLM surrogate models

In section 2, the assumption of location information will be further exemplified, while methodologies of both SBMLMs and IBMLMs will be discussed. In subsection 2.1, a half-symmetric metal forming case will be discussed to further illustrate the effect of location information. In subsection 2.2, the multi-layer perceptron (MLP) will be introduced as a typical method of SBMLMs. In subsection 2.3, the CNN will be introduced as a typical method of IBMLMs. Particularly, CNNs enhanced by multiple blocks, namely Res-SE-U-Nets, will be discussed and then utilized to develop the IBMLM model in this study.

## 2.1 Location information of FEA result

To intuitively and quatitatively explain the location information, both SBMLM and IBMLM were used to predict the max equivalent plastic strain (Max. PEEQ) of a half-symmetric metal forming simulation case with two separated binders. The settings of this case are shown in Fig 2 (a). In the Iso view, a string of constraints denotes the symmetric surface of this half model. The stamping direction is along the negative direction of z axis, while the stamping velocity is set to 100mm/s. In the top view, two separated binders with different binder forces, scilicet F1 and F2, are illustrated. CRDQ (Cold rolled, drawing quality) steel was used as the blank material, while the punch, die and binders were all set rigid. Based on these settings, plastic strain fields can be calculated, the maximum value of which are aliased as Max. PEEQ.

In this case, the binder force vector, [F1, F2], are regarded as design parameters. As shown in Fig 2 (c), there are only two samples in training set, the first of which is [1, 5] while the second is [5, 1]. Their ground truth (GT) Max. PEEQ results are both 0.7302 since the two binders are symmetric to their boundary. In test set, there is only one sample. The binder force vector of this test sample is [3, 3] which is the average of those of training samples. The GT Max. PEEQ of the test sample is 0.6348, which is lower than the average of those of training samples. The GT strain fields (GT SF) of all three samples are presented.

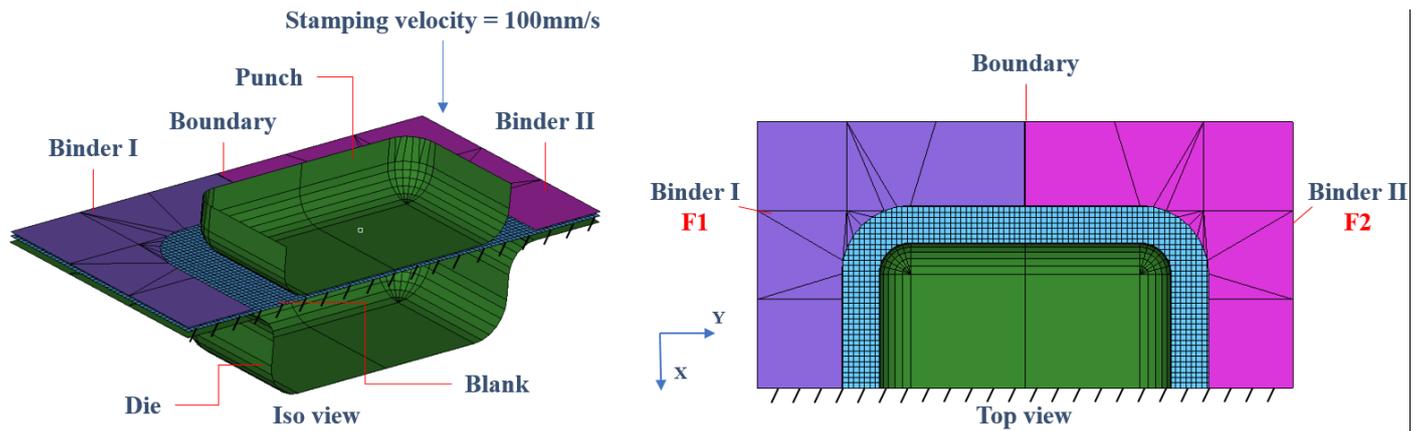

(a)

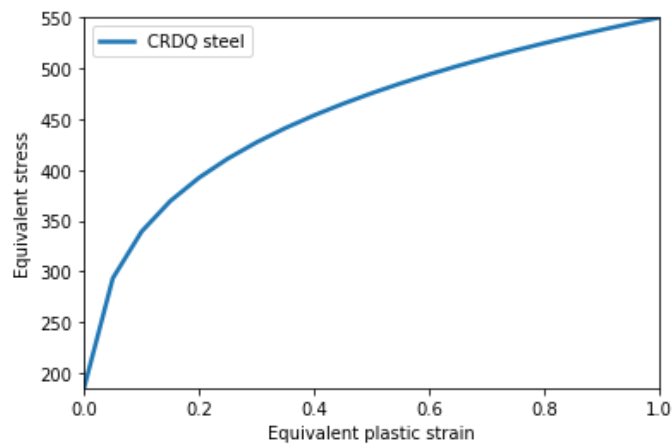

(b)

| Number | Set | F1 | F2 | Max. PEEQ | GT SF |
|---|---|---|---|---|---|
| S1 | training | 1 Mpa | 5 Mpa | 0.7302 | |
| S2 | training | 5 Mpa | 1 Mpa | 0.7302 | |
| S3 | test | 3 Mpa | 3 Mpa | 0.6348 | |

(c)

Fig 2. A surrogate model determined by two training samples to illustrate the location information of FEA results:

(a) Setup of a half-symmetric model with varying binder forces (b) Stress-Strain curve of the material used in the simulation (c) Two training samples and one test sample

F1 and F2 refers to the binder pressure of binder I and binder II; Max. PEEQ refers to the maximum value of equivalent plastic strain; GT SF refers to the ground truth stress fields as images

Based on the two-sample training set, both SBMLM and IBMLM were leveraged to predict the Max. PEEQ of the test sample. Considering prior knowledge about the mapping relationship was unknown beforehand, a linear prior was given to the SBMLM and IBMLM models. Subsequently, the three GT Max. PEEQ and two predicted Max. PEEQ can be plotted as a 3D graph, where the two independent variables were [F1, F2] while the dependent variable was Max. PEEQ. For the sake of visualization, in Fig 3 (a), this 3D graph is projected onto a 2D graph by equivalently replacing [F1, F2] with (F1-F2), which presents a plot of Max. PEEQ versus (F1-F2). In this figure, the two green dots refer to the training samples. The dark red dot refers to the GT of the test sample. The medium red dot refers to the prediction of IBMLM. The light red dot refers to the prediction of SBMLM. To illustrate, the GT SF or predicted SF (PD SF) of three samples (S1, S2, S3), if available, are placed aside each dot. It is observed that IBMLM remarkably outperforms SBMLM in accuracy and clarity. Comparing the GT fields and PD fields in Fig 3 (b), common patterns have been observed at most locations apart from the tails at the regions in red circles.

The assumption of location information can explain this discrepancy between SBMLM and IBMLM. The Max. PEEQ values of two training samples are equal in quantity while differ in locations. SBMLM can give only the average of Max. PEEQ values based on linear prior, which leads to an absolute error of 0.0954 and a relative error of 15.02%. IBMLM, on the contrary, calculate the average of plastic strain at every pixel to generate the PD SF. The Max. PEEQ of PD SF is taken as the PD Max. PEEQ. By counting in the pixel-wise location information, IBMLM reduces the absolute error to 0.0490 while the relative error to 7.719%. To elaborate, IBMLM convert the mapping relationship of the surrogate models to bijective functions by increasing the dimensionalities of inputs and outputs. This bijective mapping relationship is easier to learn.

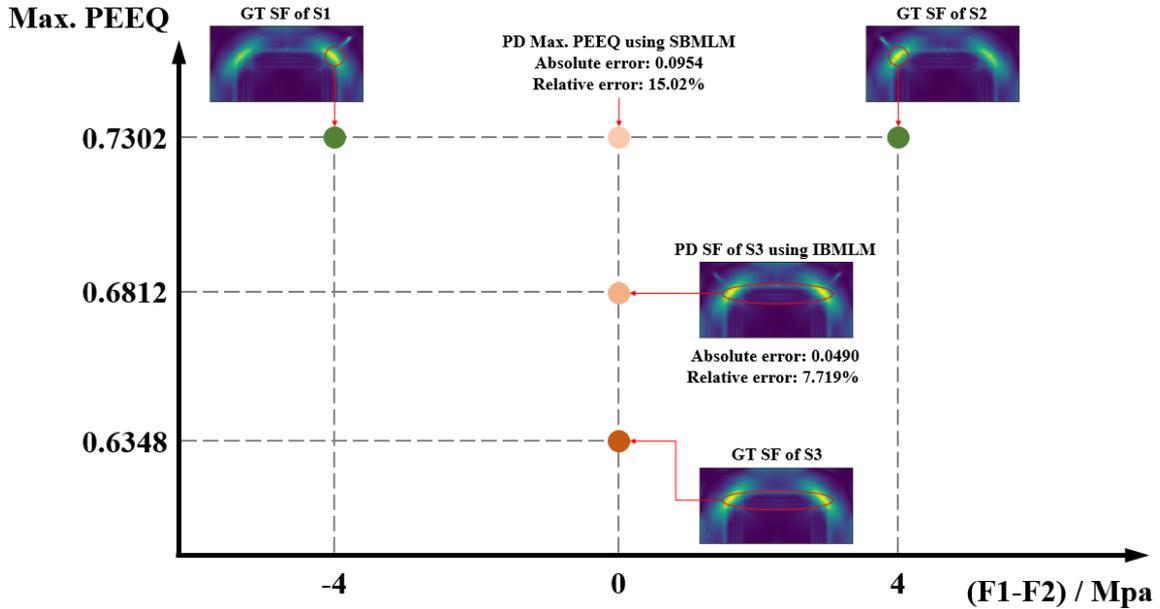

(a)

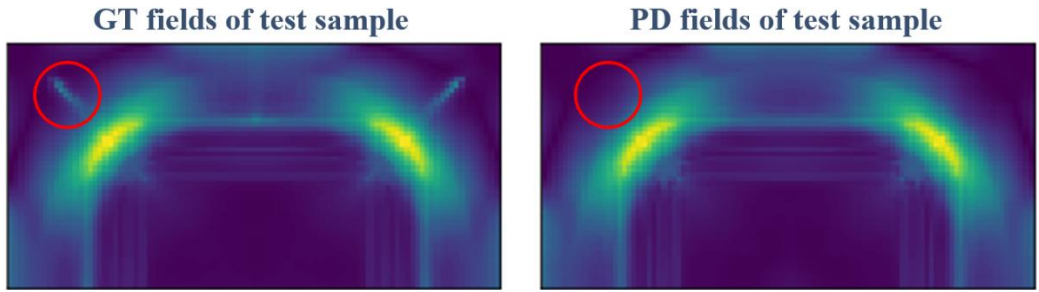

(b)

Fig 3. Max. PEEQ predictions of SBMLM and IBMLM both under linear assumption:

(a) Plot of Max. PEEQ versus (F1-F2) (b) Comparison between the GT and the IBMLM PD of test sample

One thing to notice is that this is a linear problem on a two-sample dataset. With more samples or nonlinear surrogate models, the advantages of IBMLMs will be even more prominent. The assumption of location information is thus verified. In the remainders of Section 2, specific techniques of SBMLM and IBMLM will be discussed.

**2.2 MLP for SBMLM**

As a comparison to IBMLM, in this paper, an SBMLM surrogate model will be established using MLPs. An MLP, in a narrow sense, is a typical structure of feedforward artificial neural networks (ANNs), which is comprised of an input layer, an output layer and multiple hidden layers. The neurons of two abutting layers are fully connected, where each connection is assigned with a weight. Fig 4 presents an MLP consisting of L hidden layers. In the $i^{th}$ layer ($i = 1, 2, \dots, L$), there are $ni$ neurons. In feedforward, the value of each neuron is calculated as:

$$a_j^i = A\left(\sum_{k=1}^{ni}(\omega_k^j a_k^{i-1} + b^{i-1})\right), i = 1,2,\ldots,L+1; j = 1,2,\ldots,ni$$

where the superscript refers to the layer number and the subscript refers to the neuron number in a certain layer. To clarify, a superscript of 0 refers to the input layer, while a superscript of L+1 refers to the output layer. $a_j^i$ refers to the value of the $j^{th}$ neuron in the $i^{th}$ layer. Specially, $a_1^{L+1}$ equals to the output $\hat{Y}$. $\omega_k^j$ refers to the weight of the connection from the $k^{th}$ neuron in the $(i-1)^{th}$ layer to the $j^{th}$ neuron in the $i^{th}$ layer. $b^i$ refers to the bias term in the $i^{th}$ layer. $A(\cdot)$ refers to an activation function.

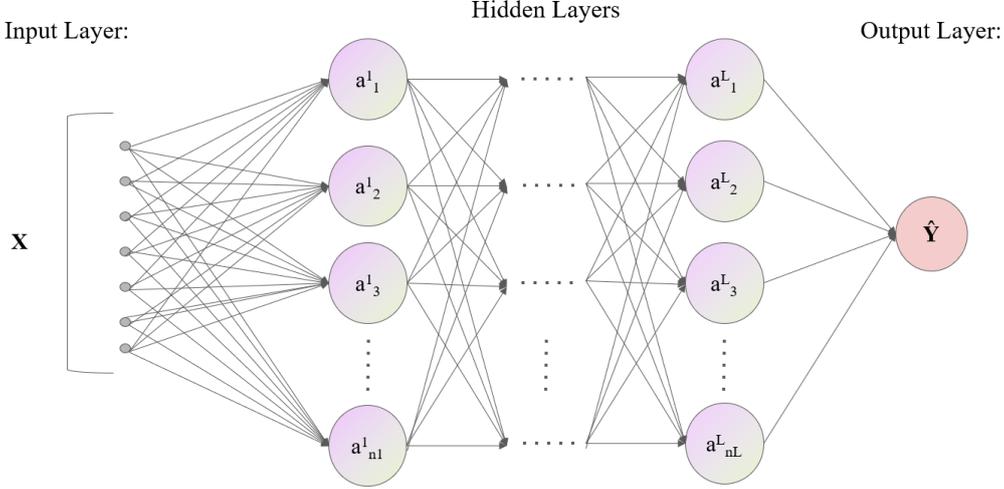

Fig 4. The schematic of an MLP for SBMLM

## 2.3 CNN for IBMLM

To establish an IBMLM surrogate model, CNNs are utilized in this paper. A major strength of CNNs is that they can hierarchically extract location information, which makes CNNs suitable for IBMLM (M. A. Islam, S. Jia, N. D. B. Bruce, 2020). Since 2014, multiple architectures of modern deep CNNs have been proposed (A. Krizhevsky, I. Sutskever, G. E. Hinton, 2012; K. He, X. Zhang, S. Ren, J. Sun, 2015; S. Ren, K. He, R. Girshick, J. Sun, 2016; K. He, G. Gkioxari, P. Dollar, R. Girshick, 2018). Among these architectures, the U-Net has been proven effective in image-to-image generating tasks like IBMLM surrogate models (O. Ronneberger, P. Fischer, T. Brox, 2015). A typical U-Net, of which the remarkable feature is the skip connection, is presented in Fig 5. U-Nets have been investigated in several IBMLM studies. N. Thuerey et. al. developed a U-Net model to fast predict the results of Reynolds-averaged Navier-Stokes simulations (2019). This model was further enhanced by physics-driven methods and applied to predicting steady heat conduction simulations (H. Ma, X. Hu, Y. Zhang, N. Thuerey, O. J. Haidn, 2020). Other IBMLM studies using U-Nets are not stated at length in this paper (S. Fotiadis et al., 2020; M. Tang, Y. Liu, L. J. Durlofsky, 2020; J. Chen, J. Viquerat, E. Hachem, 2019).

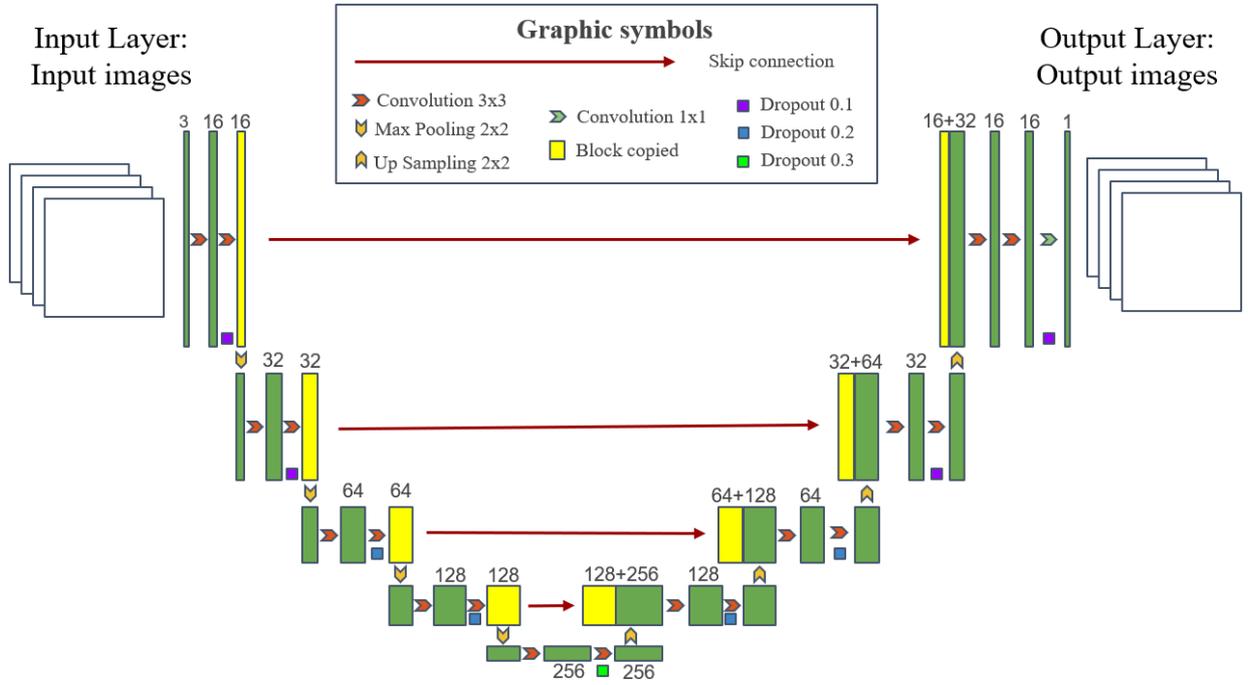

Fig 5. The schematic of a U-Net and skip connection mechanisms

To further enhance the predicting capability of U-Nets, residual connections and attention mechanisms are combined to U-Nets, which leads to Res-SE-U-Nets as shown in Fig 5 (a). To elaborate, the input $x$ of a Res-SE-U-Net is firstly processed by two integrated CNN layers into $u \in R^{H \times W \times C}$. An integrated CNN layer refers to a CNN layer that followed by a batch normalization layer (BN) and a ReLU layer. A squeeze-excitation (SE) block is then attached after the two integrated CNN layers. The specific architecture of SE blocks is shown in Fig 5 (b). In an SE block, the input tensors $u$ are reduced to $w \in R^C$ by a global pooling layer. $w$ is downscaled to $p \in R^{C/r}$ by a fully connected layer (FC+ReLU), where $r$ refers to the reduction ratio of a SE block. $p$ is then upscaled to $q \in R^C$ by another fully connected layer (FC+Sigmoid). $q$ is reshaped to $e \in R^{H \times W \times C}$ that has the same dimensions as $u$. The SE block output $v$ is then calculated by adding $u$ to $e$. $v$ is then processed by the activation function $F(\cdot)$ in Fig 5 (a). The Res-SE block output $v$ is calculated by adding $x$ to $F(v)$.

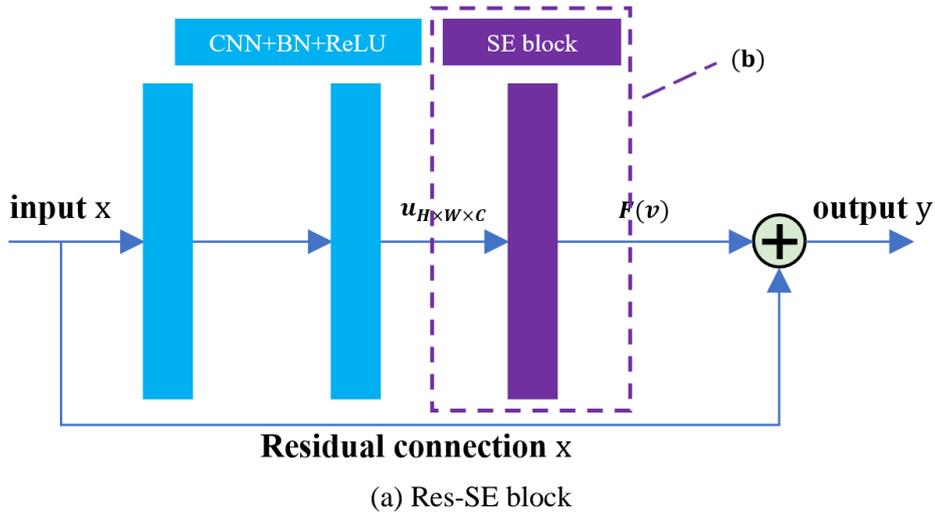

(a) Res-SE block

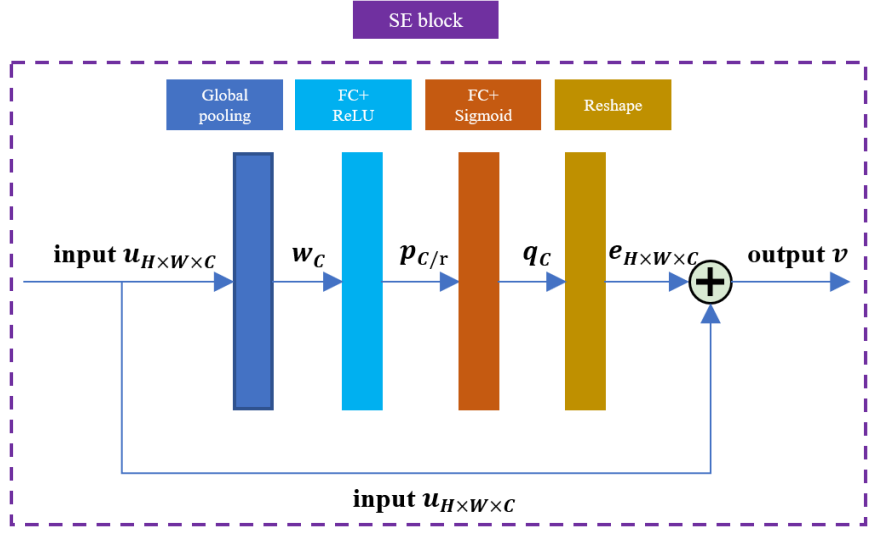

(b) SE block

Fig 6. Composition of Res-SE block

The Res-SE blocks discussed above have been combined to U-Nets in several studies (Z. Nie, 2020; Z. Nie, T. Lin, H. Jiang, L. B. Kara, 2020; H. Jiang, Z. Nie, R. Yeo, A. B. Farimani, 2020). A remarkable increase of accuracy has been observed using these Res-SE-U-Nets. Therefore, in this present paper, a Res-SE-U-Net will be developed to establish IBMLM models.

## 3. Dataset & Specific surrogate models

To verify the location information assumption discussed in section 2, specific SBMLM and IBMLM surrogate models are developed and compared based on a dataset of metal forming simulations. This dataset was composed of $N$ pairs of design parameters and simulation results, where $N$ referred to the number of samples. Subsequently, the dataset was divided into a training set and a test set. An MLP model was developed as an SBMLM surrogate model, which predicts the scalar based simulation results, namely the Max. PEEQ, from the scalar based design parameters; a Res-SE-U-Net model was developed as an IBMLM surrogate model, which maps from the image based design parameters to the image based simulation results, scilicet the plastic strain fields.

For the sake of comparability, activation functions, optimizers, loss metrics and regularization terms were unified for both the MLP and Res-SE-U-Net. The rectified linear unit (ReLU) was selected as the activation function (V. Nair, G. E. Hinton, 2010). The Adam algorithm was used to train both models (D. P. Kingma, J. L. Ba, 2014). Two loss metrics were selected, namely the mean square error (MSE) and the maximum plastic strain error (MPE):

$$MSE = \frac{1}{n}\sum_{i=1}^{n}(p_i^{PD} - p_i^{GT})^2$$
$$MPE = \max(p^{PD}) - \max(p^{GT})$$

where the n refers to the output number; $p$ refers to the output vector while $p_i$ refers to the $i^{th}$ component of $p$; A superscript of PD refers to the predicted value while GT refers to the ground truth value. Particularly in the Res-SE-U-Net, $p$ refers to the output image of plastic strain fields while $p_i$ refers to the $i^{th}$ pixel of the image. To clarify, MSE was used as the loss function in the training processes of both models; MLP was used to evaluate the capability of both models to predict the Max. PEEQ. To highlight the impact of location information, regularization terms are not included. This is because regularization is purely a non-interpretable numerical trick.

In subsection 3.1, the dataset will be elaborated, based on which the interpolation task and the extrapolation task are defined. In subsection 3.2, the architectures and parameters of the MLP will be demonstrated. In subsection 3.3, the architectures and parameters of the Res-SE-U-Net will be demonstrated.

### 3.1 FEA dataset of a simplistic metal forming case

To conveniently elaborate the concept of IBMLMs, in this study, a simplistic dataset of cold steel stamping simulations was established, which comprised 1,080 samples. Once the IBMLMs are validated on this dataset, they can be extended to more practical and complex scenarios of metal forming. The setup of this dataset is shown in Fig 6, where the left figure shows the designed geometry and the right figure shows the quarter FEA model in Abaqus. A designed geometry was decided by five scalar variables that highlighted in green, namely $R1$, $R2$, $R3$, $BF$, and $T$. To clarify, the first three variables referred to the three radii that decided designed geometries; The $BF$ referred to binder forces; $T$ referred to thickness of blanks. Other geometric parameters, including $C1$, $C2$, $C3$ and the blank size, were fixed. The values of the first three constants were $25mm$, $40mm$, $40mm$, respectively. The blank was a square sized $70mm \times 70mm$. In the FEA model, the blank was meshed to 2500 square pixels sized 1.4mm×1.4mm.

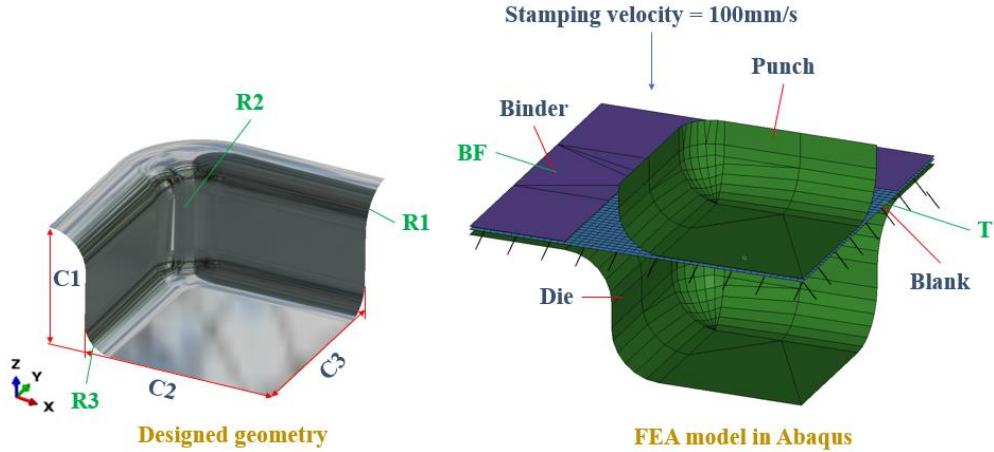

Fig 7. Setup of a quarter-model metal forming simulation: Design geometry and FEA model in Abaqus

The value of each radius, namely $R1$, $R2$, or $R3$, took the range of $[6mm, 8mm, 10mm]$. This led to 27 types of designed geometries in total by fully combining the values of $R1$, $R2$, and $R3$. For brevity, these 27 designed geometries were uniformly numbered from $Geo1$ to $Geo27$. To clarify, $Geo1$ referred to $R1 = 6mm$, $R2 = 6mm$, $R3 = 6mm$; $Geo2$ referred to $R1 = 6mm$, $R2 = 6mm$, $R3 = 8mm$, and so on. In the FEA model of each $Geo$, the $BF$ could take on any of the 20 values from $0.25Mpa$ to $5Mpa$ with a step of $0.25Mpa$; The $T$ could take on a value from $[1mm, 1.5mm]$. In total, this dataset contained $27 \times 20 \times 2 = 1,080$ samples. The unshuffled sequence of these samples is presented in Table 1.

The samples in this dataset were shuffled and then divided into a training set and a test set by a test ratio of 10%, which was regarded as an interpolation task. In this interpolation task, the test set was uniformly sampled from the whole dataset. Besides the interpolation task, an extrapolation task was defined in this study by directly dividing the samples in the unshuffled order. In this extrapolation task, parts of the geometries, scilicet $Geo26$ and $Geo27$, were never observed in the training set. Therefore, this extrapolation task could thoroughly test the generalizability and information extraction capability of SBMLM and IBMLM models.

Table 1. Dataset of the quarter-model metal forming simulation

| Geo | Geo1 | | | | ... | Geo25 | | | | ... | Geo27 | |
|---|---|---|---|---|---|---|---|---|---|---|---|---|
| BF | BF1 | ... | BF20 | | ... | BF6 | | BF7 | | ... | ... | BF20 |
| T | T1 | T2 | ... | T1 | T2 | ... | T1 | T2 | T1 | T2 | ... | T1 | T2 |

Geo: Geometry from combination 1 to 27 (27 in total)  
BF: Pressure from 0.25Mpa to 5.0Mpa (20 in total)  
T: Thickness of 1mm and 1.5mm (2 in total)

Note: 90% samples in the training set while 10% in the test set  
The dataset in initial order leads to extrapolation.  
The dataset in uniformly shuffled order leads to interpolation.

One thing to note is that the dataset size was intentionally set to a small number of 1,080 compared with that in other studies (C. Zimmerling, 2019; Z. Nie, 2020). The purpose of this small dataset was to claim that IBMLMs were superior in extracting more information from a given dataset, while not reliant on big data. To further illustrate the concept of location information, four simulation results with different $BF$ are presented in Fig 8. It can be observed that wrinkles occur under a low binder force while thickening occurs under a high binder force. These simulation results from left to right show an evolution of spatial features that contain the location information. This observation qualitatively demonstrates the necessity of using IBMLM models.

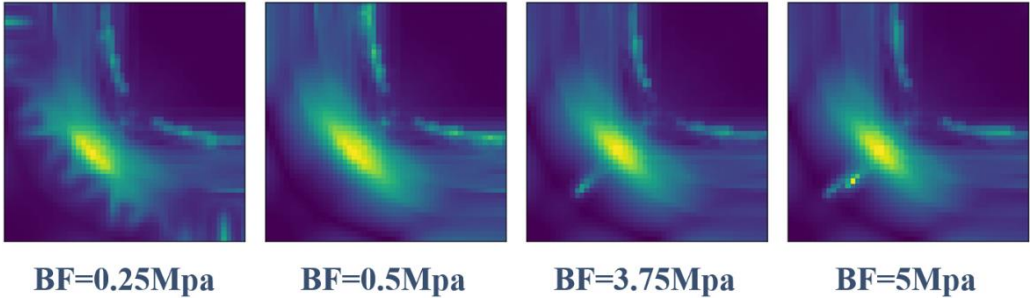

Fig 8. Evolutions of the underlying distributions of GT simulation results: Evolution along with binder forces

**3.2 The architectures and parameters of the MLP**

In the present study, to provide a reference for the IBMLM surrogate model, an MLP was developed to establish an SBMLM surrogate model. This MLP contained 6 hidden layers, the architectures and parameters of which are illustrated in Fig 9. The inputs of the MLP were scalar based design parameters, namely, $Geo$, $BF$, $T$. The output of the MLP was the Max. PEEQ. Other settings of the MLP, including activation functions, optimizers, loss metrics and regularization terms, complied with those discussed at the beginning of section 3.

Notably, compared with CNNs, MLP models are more vulnerable to gradient problems, which lead to an unfair comparison between the MLP SBMLM and the Res-SE-U-Net IBMLM. To make this comparison more credible, the layer number of this MLP was fixed to 6 while the neuron number parameters of this MLP model were well optimized. In other words, the model in Fig 9 was able to reflect the average learning capability of MLPs.

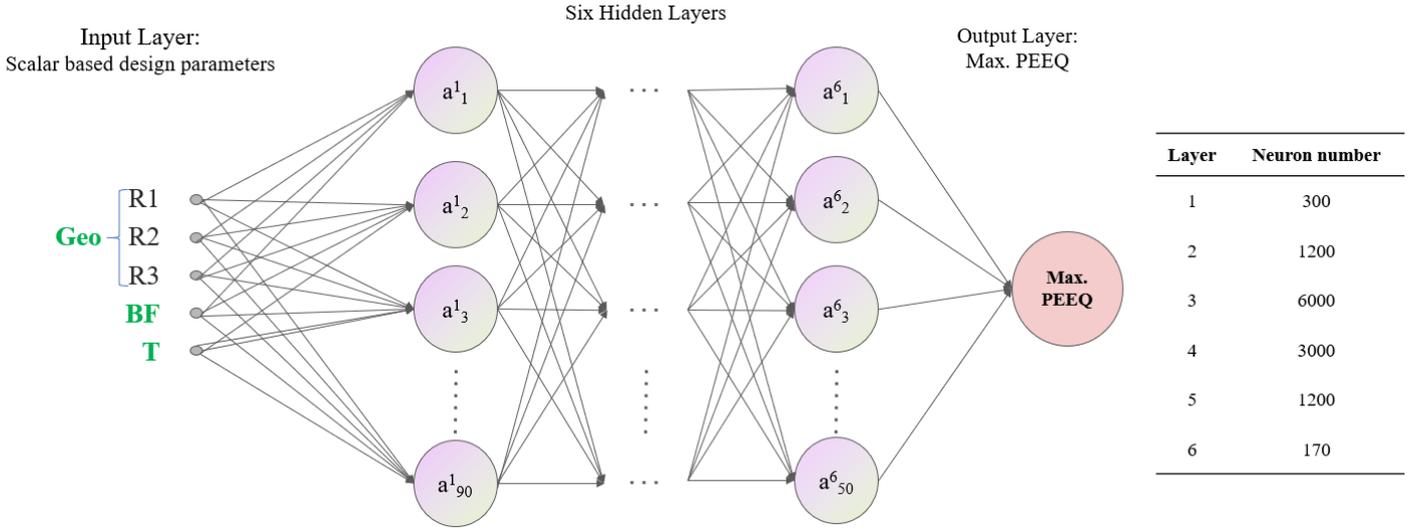

Fig 9. The architectures and parameters of the MLP used in this study

**3.3 The architectures and parameters of the Res-SE-U-Net**

To represent the design parameters using IBMLM, the three design parameters of $Geo$, $BF$ and $T$ were projected to the blank plane sized 70mm×70mm and then discretized into $199^2$ square grids. This projection led to inputs sized $199^2 \times 3$, while each input contained 3 channels of input images. To represent the simulation results using IBMLM, the plastic strain fields were plotted on the undeformed blank that discretized into $50^2$ pixels. This plot led to outputs sized $50^2 \times 1$. Based on these IBMLM inputs and outputs, a Res-SE-U-Net was developed to establish an IBMLM surrogate model. This Res-SE-U-Net was composed of three modules, namely downscale, upscale, and Res-SE blocks. In the downscale module, the ground truth IBMLM (GT IBMLM) design parameters were gradually reduced to feature images sized $12^2 \times 512$ by a series of CNN layers. Meanwhile, the channel number was gradually increased from 3 to 512, which aimed to extract the location information of the design parameters. These feature images were then processed by 6 serial Res-SE blocks after the downscale module and then imported to the upscale module. In the upscale module, the feature images sized $12^2$ were expanded by a series of transpose CNN layers to predicted IBMLM (PD IBMLM) simulation results sized $50^2$. Particularly, skip connections were established between the downscale module and the upscale module to enhance the capability of information extraction.

The architecture of this Res-SE-U-Net is illustrated in Fig 10 while the parameters of CNN and transpose CNN layers in this network are presented in Table 2. Other settings of the MLP, including activation functions, optimizers, loss metrics and regularization terms, complied with those discussed at the beginning of section 3.

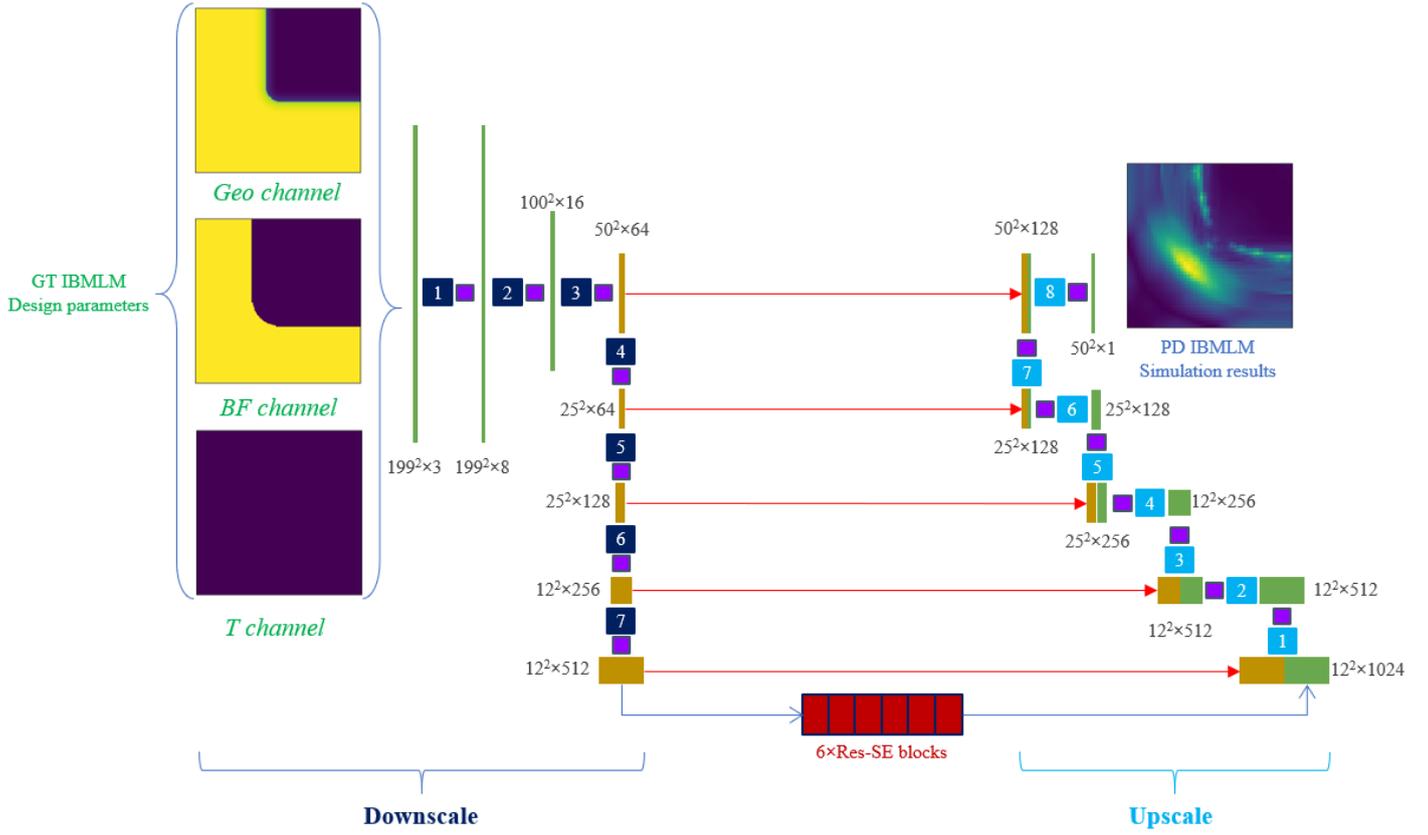

Fig 10. The architectures of the Res-SE-U-Net used in this study

Table 2. Parameters of the Res-SE-U-Net

| Layer Number | Downscale | | | Upscale | | | Res-blocks |
|---|---|---|---|---|---|---|---|
| | Kernel | Stride | Padding | Kernel | Stride | Padding | |
| 1 | (1,1) | (1,1) | (0,0) | (3,3) | (1,1) | (1,1) | Number of blocks: 6 |
| 2 | (11,11) | (2,2) | (5,5) | (3,3) | (1,1) | (1,1) | |
| 3 | (8,8) | (2,2) | (3,3) | (3,3) | (1,1) | (1,1) | CNN layers in each block: Kernel(3,3), Stride(1,1), Padding(1,1) |
| 4 | (6,6) | (2,2) | (2,2) | (3,3) | (2,2) | (0,0) | |
| 5 | (3,3) | (1,1) | (1,1) | (3,3) | (1,1) | (1,1) | |
| 6 | (3,3) | (2,2) | (0,0) | (3,3) | (1,1) | (1,1) | |
| 7 | (3,3) | (1,1) | (1,1) | (6,6) | (2,2) | (2,2) | Reduction ratio in SE blocks: r = 16 |
| 8 | | | | (3,3) | (1,1) | (1,1) | |

## 4. Results & Discussions

The Max. PEEQ values of the dataset range from 0.6366 to 0.9639. Based on this value range and engineering experiences, the maximum acceptable MPE (MCMPE) is set to be 0.04 in this study. One thing to notice that this PEEQ range is impractical since damage will have occurred when PEEQ is higher than around 0.5. However, this dataset can still be used to validate IBMLMs since the prediction capability of IBMLMs are relatively independent from the value range of the dataset. In addition, this dataset without damage models is less time-intense and contains distinct spatial features of plastic strain fields, which is beneficial for a quick validation of IBMLMs. Therefore, this dataset was used to validate IBMLMs, while the validated IBMLMs will later be extended to practical datasets of hot stamping processes for Boron steels.

The MLP SBMLM model and Res-SE-U-Net IBMLM model were evaluated respectively on both the interpolation task and the extrapolation task. To evaluate the accuracy and generalizability, four scalar indicators are applied: MPE on training set, MPE on test set, MSE on train set, MSE on test set. These scalar indicators were calculated respectively for both models and then compared. To evaluate the capability of the IBMLM model to extract location information, a fifth indicator of PWE is applied. PWE is an image indicator that can visualize the difference between the IBMLM predicted strain fields and the ground truth.

All codes were written in Pytorch. Both models ran on an NVIDIA Quadro RTX 5000 GPU. The epoch number on the interpolation task was 8,000 while that on the extrapolation task was 4,000. For both models, the thread number was set to 4 while the batch size was set to 54. Evaluation results on the interpolation task are be discussed in subsection 4.1, while those on the extrapolation task are be discussed in subsection 4.2. The advantages of IBMLM models in information extraction are demonstrated in subsection 4.3.

**4.1 Evaluation results on the interpolation task**

An interpolation task was designed to certify whether IBMLM was able to predict the Max. PEEQ, which is evaluated by MPE. One thing to notice is that, MPE is explicitly included in the loss function of the MLP SBMLM model while the loss function of the Res-SE-U-Net IBMLM model doesn't explicitly contains MPE. It is thus anticipated that the MLP model should perform well, while the performance of the Res-SE-U-Net model is not guaranteed. Therefore, this task can be used to certify whether the Res-SE-U-Net model really extract inherent information that guarantees the MPE accuracy.

The results of the four scalar indicators on the interpolation task are illustrated in Fig 11 (a) (b) (c) (d). In each figure, one of the four indicators are compared between the MLP SBMLM model and the Res-SE-U-Net IBMLM. To present informative figures, the MPEs of both models are sorted in ascending order, respectively on the training set and test set.

In Fig 11 (a) (b), both SBMLM and IBMLM performs well since all MPE errors are under MCMPE. In Fig 11 (b), IBMLM slightly outperformed SBMLM. This manifests a better generalizability of IBMLM over SBMLM in the interpolation task. In Fig 11 (c), the curves of logarithmic MSE (LMSE) on training set are plotted for both models. Both curves decrease to the same order of magnitude of -6 after 5000 epochs. However, the fluctuation amplitude of the SBMLM curve is significantly larger than that of the IBMLM curve. In Fig 11 (d), the curves of LMSE on test set are plotted for both models. The IBMLM test error converges after 1500 epochs whereas the SBMLM curve still acutely fluctuates after 8000 epochs. Fig 11 (c)(d) manifest a better convergency and robustness of IBMLM over SBMLM, which are of vast importance in practical applications. This advantage of convergency and robustness is because that IBMLM converts the surrogate model to an easy-to-train bijective function.

To conclude, both models performs well on the interpolation task, whilst IBMLM performs better in generalizability, robustness and convergency. As mentioned above, given an interpolation task tailored for SBMLM, IBMLM still slightly outperforms SBMLM. This indicates that IBMLM models indeed extract and predict the inherent features of simulation results.

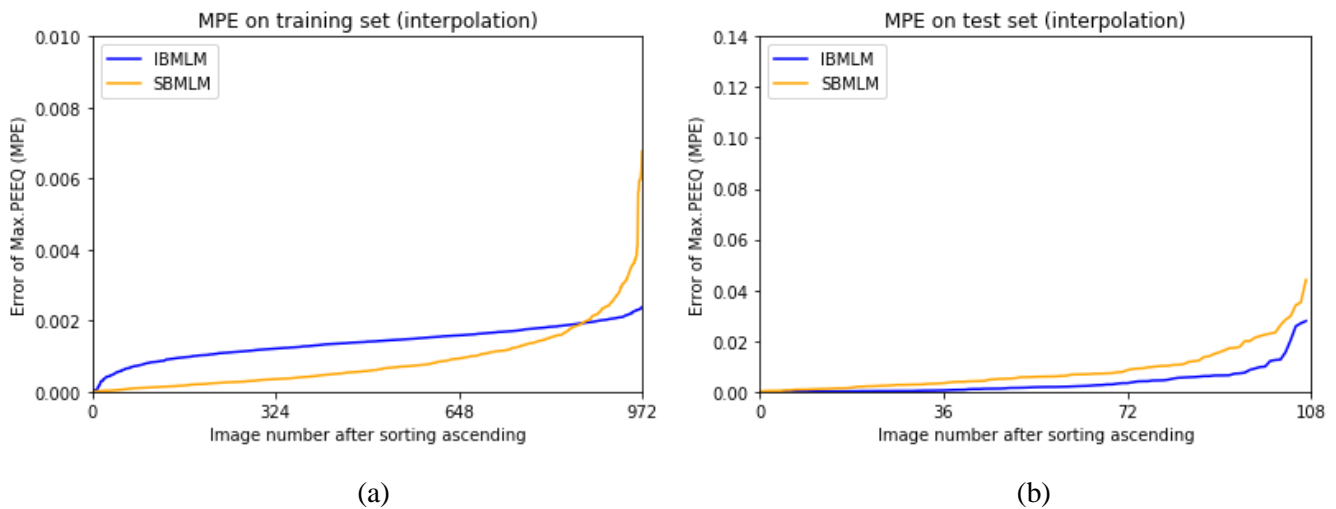

(a) (b)

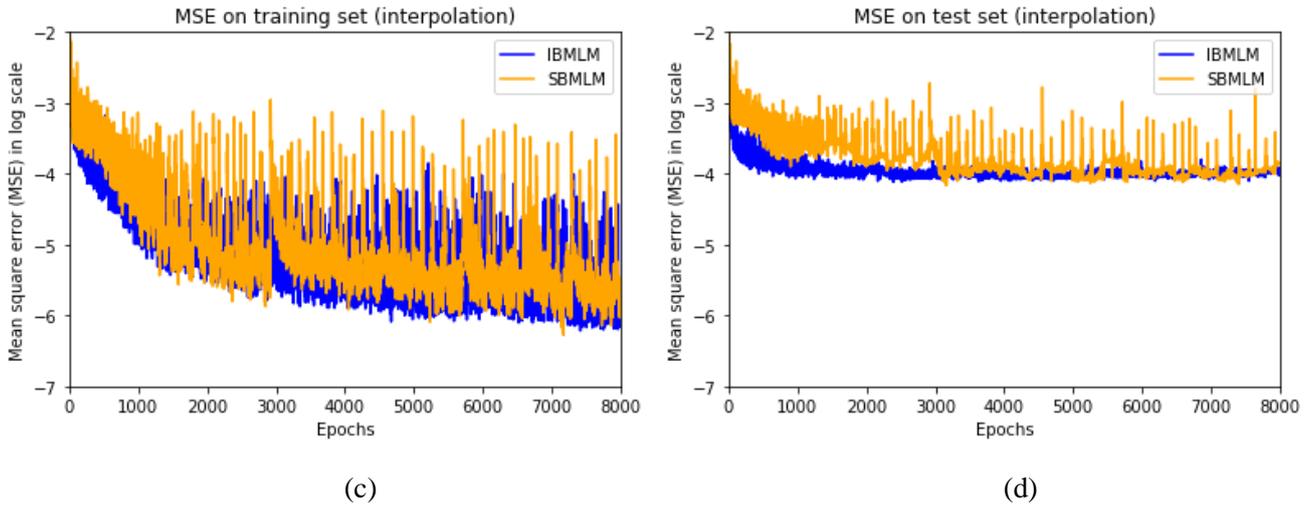

(c)           (d)

Fig 11. Interpolation performance of SBMLM and IBMLM: (a) MPE on training set (b) MPE on test set (c) MSE loss on training set (d) MSE loss on test set

**4.2 Evaluation results on the extrapolation task**

As extrapolation effects are ineluctable in practical industrials, an extrapolation task, in which the test geometries were unseen, was designed to test the generalizability of both models.

Analogously, the results of the four scalar indicators on the extrapolation task are presented in Fig 12 (a) (b) (c) (d). In Fig 12 (a), the training set performances of both models on the extrapolation task are as good as those on the interpolation task. In Fig 12 (b), IBMLM remarkably outperforms SBMLM on the test set of the extrapolation task, while SBMLM is unacceptably overfitted. This is because the $Geo$ parameters in the test set are completely unseen in the training set. In this case, SBMLM fails to generalize because of the lack of location information in both scalar based inputs and outputs. IBMLM, on the contrary, has a certain degree of generalizability since it preserves location information. In Fig 12 (c), both curves of IBMLM and SBMLM converge to the same order of magnitude, whereas the convergency of the latter one results from overfitting. In Fig 12 (d), the IBMLM is stabilized around -3.7 after 1000 epochs, whereas the SBMLM still slightly oscillate around -3.2 after 4000 epochs.

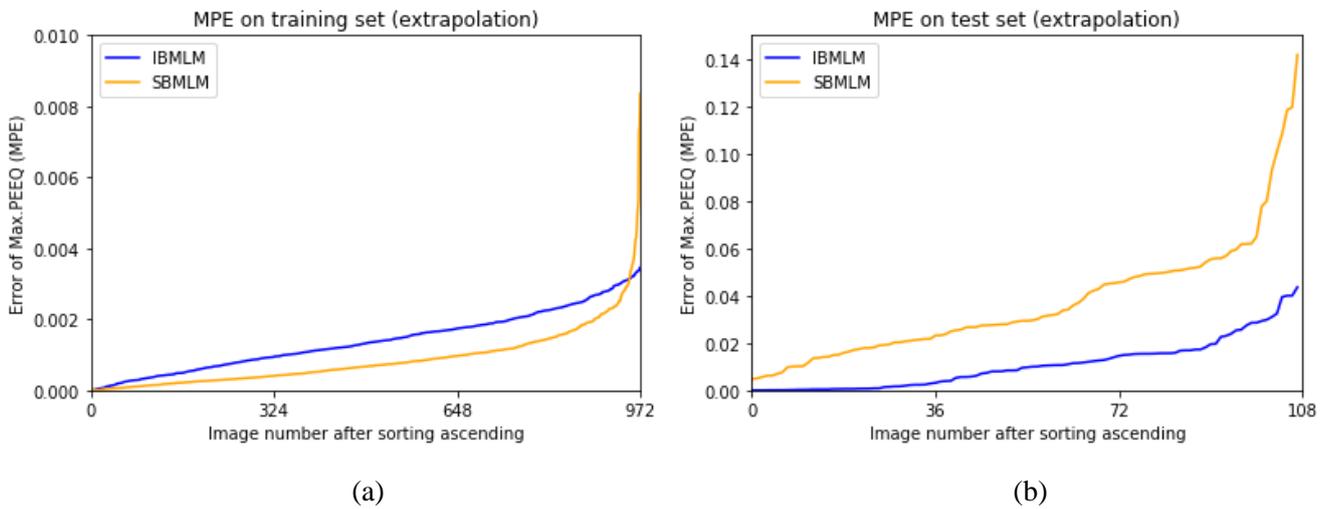

(a)           (b)

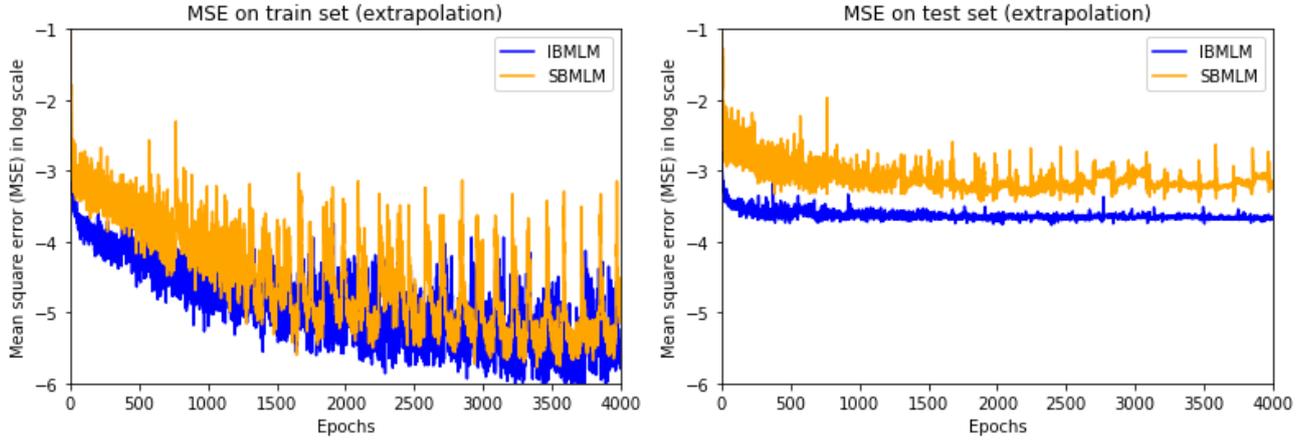

Fig 12. Extrapolation performance of SBMLM and IBMLM (a) MPE on training set (b) MPE on test set

To further stress the importance of evaluations on the extrapolation task, the comparison of test set performances on both tasks are presented in Fig 13 (a) (b). It is observed that both SBMLM and IBMLM perform better on the interpolation task. This is because that the training set in the interpolation task covers the ranges of design parameters better than that does in the extrapolation task. To better cover the ranges of design parameters, advanced sampling strategies are adopted, which are not elaborated in detail in this paper (M. Stein, 1987). Despite these strategies, it is impossible to guarantee a perfect sampling and eradicate the effects of extrapolation. This problem of extrapolation effects is extremely severe when confronting complicated and non-isometric geometries (R. Marin et. al., 2020).

Fortunately, these effects can be further mitigated by IBMLMs even without advanced sampling strategies. As shown, in Fig 13, when an extrapolation task is converted to an interpolation task using sampling strategies, the IBMLM model can obtain robust benefits from these strategies. This indicates that IBMLM models are more efficient in information extraction.

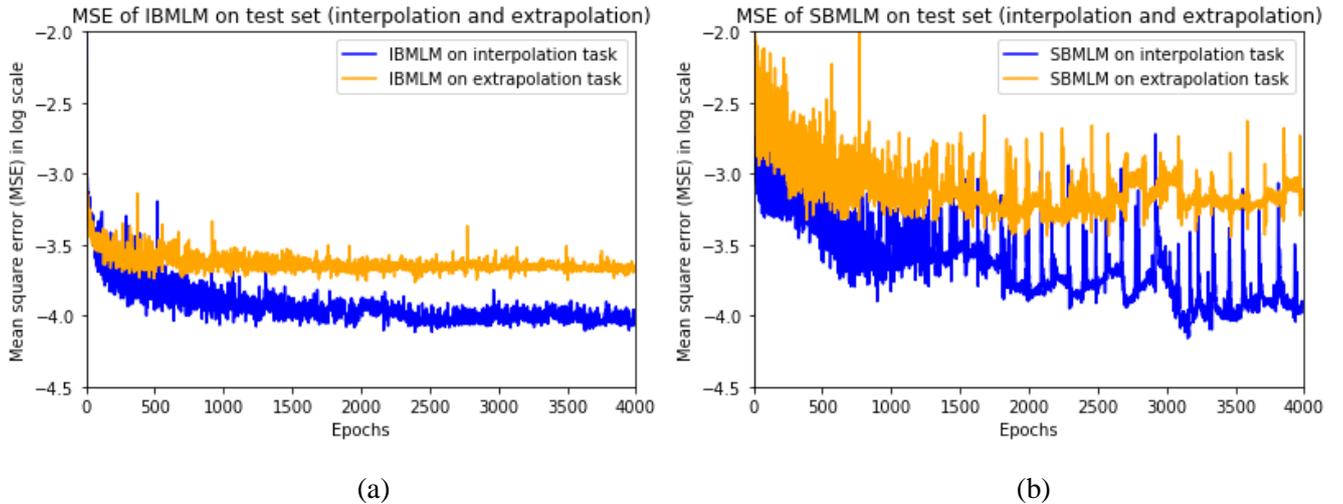

(a)          (b)

Fig 13. Comparison of test set performances on interpolation and extrapolation tasks: (a) SBMLM (b) IBMLM

### 4.3 Information advantage of IBMLM

As mentioned in the introduction, a critical advantage of IBMLM models is that they can be applied to scenarios where the outputs are definitely impossible to be represented by scalars, including thinning distribution analysis, punch impact line analysis, A-class surfaces design, microstructure distributions. To evaluate the capability of IBMLM to extract location information, pixel-wise errors (PWEs) are used, which are in the form of images. PWEs are calculated as the pixel-wise difference between a ground truth (GT) plastic strain field and its prediction (PD) by IBMLM:

$$PWE = GT\ field - PD\ field$$

The GT, PD fields and PWEs of several typical cases are visualized in Fig 14. In Fig 14 (a), 6 cases from the training set are presented. The PD fields are practically identical with the GT fields while the PWEs are low and evenly distributed. In Fig 14 (b), 6 cases from the test set are presented. It is observed that the PWEs of these test cases are higher than those of the training cases. However, the extreme values of PWEs are mostly centralized in zones of sensitive features, such as the tail downward to the left in test Case2. Mostly, these extreme PWEs slightly perturb the locations of the sensitive features instead of radically transforming the features. Therefore, these extreme PWEs are acceptable. To further reduce the extreme values, advanced sampling and large datasets might be required instead of just optimizing the surrogate models, which are not discussed in this paper.

Based on the discussions above, it is affirmed that IBMLM surrogate models are capable of extracting and visualizing the location information of simulation results. One advantage of location information is that, strange images predicted by IBMLM surrogate models can provide more information for the researchers to determine the possible reasons of errors. Strange scalars predicted by SBMLM, on the contrary, can only indicate the existence of errors, while cannot indicate the locations and reasons of errors.

One thing to note is that the magnitudes of PWEs seem high compared with those of MSEs and MPEs mentioned above. However, this comparison is pointless. This is because MSEs and MPEs are used to contrast the accuracy and generalizability between SBMLM and IBMLM; while PWEs are used to evaluate the capability of IBMLM of extracting location information. To comparatively judge the effectiveness of PWEs, advanced indicators are required in future studies.

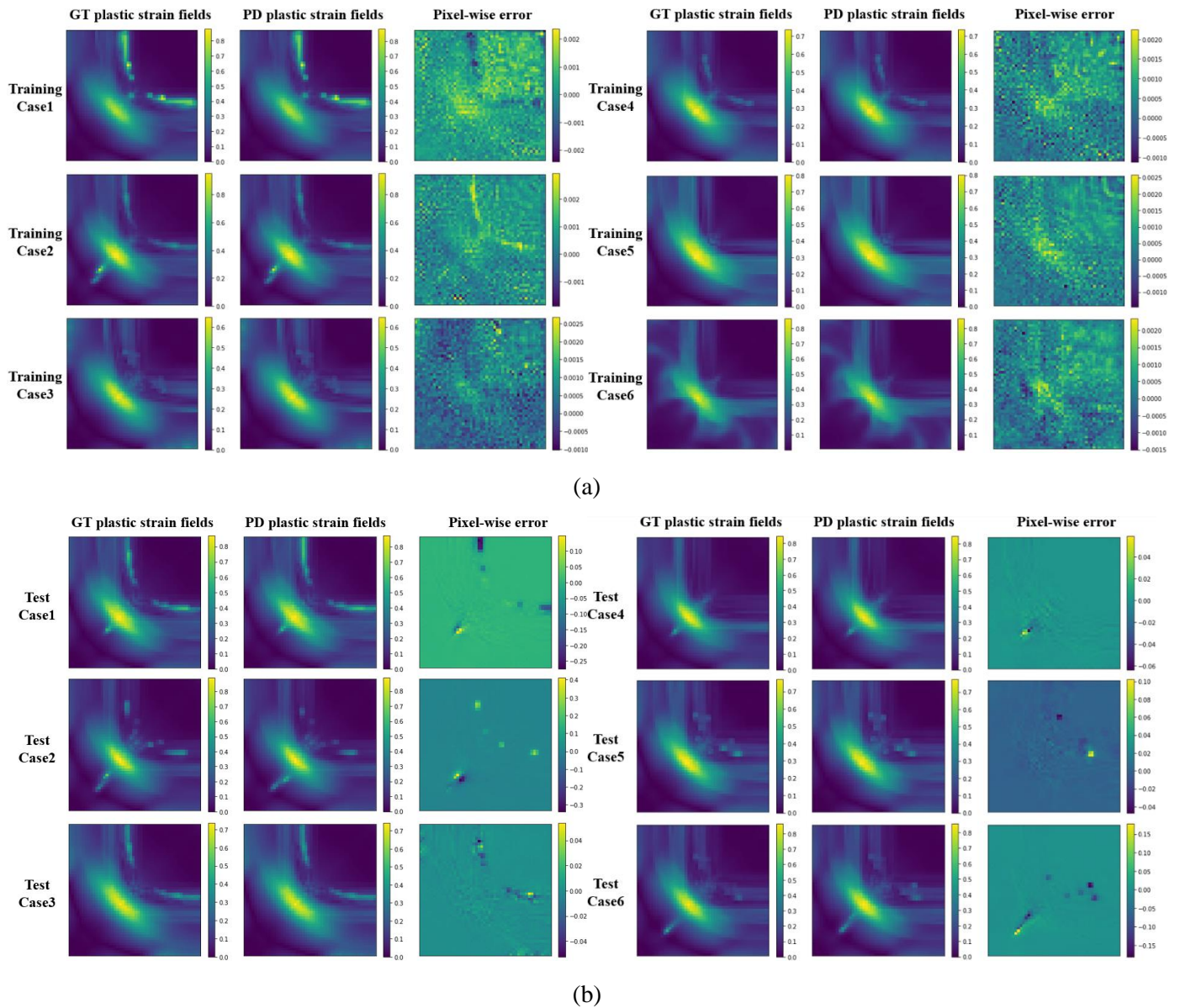

Fig 14. Spatial distributions of simulation results predicted by IBMLM surrogate models (Normalized colour maps):

(a) Cases in training set (b) Cases in test set

## 5. Conclusion & Outlook

In this study, a Res-SE-U-Net IBMLM surrogate model was innovatively applied to predicting the plastic strain fields of stamp forming simulations. Compared with a SBMLM model, this IBMLM model is more accurate, generalizable and robust on both interpolation and extrapolation tasks. To refine, the training process of the IBMLM model converged faster while the test error was remarkably reduced. In terms of generalizability, when tackling unseen design parameters, such as unseen complicated geometries, existing IBMLM models can be transferred and fine-tuned, while SBMLM models are usually discarded. This possibility of transfer learning further enhances the advantages of IBMLM models.

In terms of theories and interpretability, compared with other early studies about the IBMLM, the underlying assumption of location information was intuitively and quantitatively discussed in this study. Based on this assumption, the advantage of IBMLM is concluded as extracting more information from datasets of given sizes. Meanwhile, the output images of IBMLM provided more location information for the researchers to evaluate and accurately calibrate the surrogate model.

Despite these efforts and achievements, the IBMLM model suffered from the problem of training time. For example, in the interpolation task, a 1500-epoch training process of the IBMLM model, which was sufficient for convergency, costed 3.75 hours, while that of the SBMLM model costed 0.8 hours. This is because that IBMLM take full images as inputs and outputs, which increases the computational complexity of IBMLM. However, due to the strong generalizability, IBMLM models can cover variant geometries in a single model, while considerably reduce the sampling points. IBMLM models can be trained one-shot and then transferred to other scenarios. This considerably reduces the time spent on sampling and simulations, which make up the most majority of computation time. Therefore, IBMLM models are time saving in the long run. Moreover, it has been observed that IBMLM models are easier to be optimized and lightened. Therefore, lightweight IBMLM architectures can be designed to further mitigate the computation burden.

In future studies, IBMLM will be extensively investigated. Firstly, the design principles of architectures will be studied to boost performance as well as save training time. Secondly, multiple machine learning methods and tricks, such as generative models and physics-based prior, will be leveraged. Thirdly, IBMLM models are promising to be transferred to multiple scenarios, particularly where it is necessary to represent the simulation results by images. Furthermore, IBMLM models will be embedded in inverse problems, such as optimizing the metal forming manufacturing parameters.